\documentclass[sigconf]{acmart}
\renewcommand\footnotetextcopyrightpermission[1]{}

\usepackage{enumitem} 
\usepackage{makecell}
\usepackage{balance}

\AtBeginDocument{%
  }

\begin{document}


\title[MoveGCL]%
{Breaking Data Silos: Towards Open and Scalable Mobility\\
Foundation Models via Generative Continual Learning}

\author{Yuan~Yuan*, Yukun~Liu*, Chonghua~Han, Jie~Feng, Yong~Li}




\affiliation{
\country{Center for Urban Science and Computation, Tsinghua University}
}

\renewcommand{\shortauthors}{Yuan et al.}
\begin{abstract}
  Human mobility is a fundamental pillar of urban science and sustainability, providing critical insights into energy consumption, carbon emissions, and public health. However, the discovery of universal mobility laws is currently hindered by the ``data silo'' problem, where institutional boundaries and privacy regulations fragment the necessary large-scale datasets. In this paper, we propose MoveGCL, a transformative framework that facilitates collaborative and decentralized mobility science via generative continual learning. 
  MoveGCL enables a distributed ecosystem of data holders to jointly evolve a foundation model without compromising individual privacy. The core of MoveGCL lies in its ability to replay synthetic trajectories derived from a generative teacher and utilize a mobility-pattern-aware Mixture-of-Experts (MoE) architecture.
  This allows the model to encapsulate the unique characteristics of diverse urban structures while mitigating the risk of knowledge erosion (catastrophic forgetting). 
  With a specialized layer-wise progressive adaptation strategy, MoveGCL ensures stable convergence during the continuous integration of new urban domains. Our experiments on six global urban datasets demonstrate that MoveGCL achieves performance parity with joint training, a previously unattainable feat under siloed conditions. This work provides a scalable, privacy-preserving pathway toward Open Mobility Science, empowering researchers to address global sustainability challenges through cross-institutional AI collaboration.
  To facilitate reproducibility and future research, we have released the code and models at \color{blue}{\url{https://github.com/tsinghua-fib-lab/MoveGCL}}.
\end{abstract}

\maketitle

\renewcommand{\thefootnote}{\fnsymbol{footnote}} 
\footnotetext[1]{Equal contribution.}

\section{Introduction}

The study of human mobility is more than a technical task; it is a cornerstone of urban systems theory and sustainability science~\cite{xu2025using,huang2015defining}. Understanding how billions of people move through urban environments is essential for optimizing transportation efficiency, reducing the global carbon footprint, and managing epidemiological risks~\cite{nouvellet2021reduction,zeng2024estimating,zheng2023spatial}. In the era of Foundation Models, fields such as 
natural language processing (NLP)~\cite{kojima2022large,brown2020language,bi2024deepseek} and computer vision (CV)~\cite{esser2024scaling,liu2024sora} have undergone a paradigm shift. 
This evolution presents an unprecedented opportunity to move beyond insular empirical studies, which often focus on isolated datasets and specific cities, toward a unified, generalizable science of mobility.

However, a critical barrier remains: the fragmentation of mobility data across organizational and national boundaries~\cite{yuan2025worldmove,zhu2023synmob}. 
Unlike the public-domain text used in NLP and CV, mobility data is inherently sensitive, held by disparate telecommunications operators or transit agencies in isolated ``data silo'' due to privacy concerns and stringent regulatory frameworks (e.g., GDPR)~\cite{kong2023mobility,yuan2025learning}.
This isolation prevents the scientific community from training large-scale models that can generalize across different urban morphologies and socio-economic contexts. Consequently, the field is stuck in a paradigm where models are trained on small, localized datasets, failing to capture the complex, universal patterns of human movement.

\begin{figure*}[t]
  \centering
  \includegraphics[width=\linewidth]{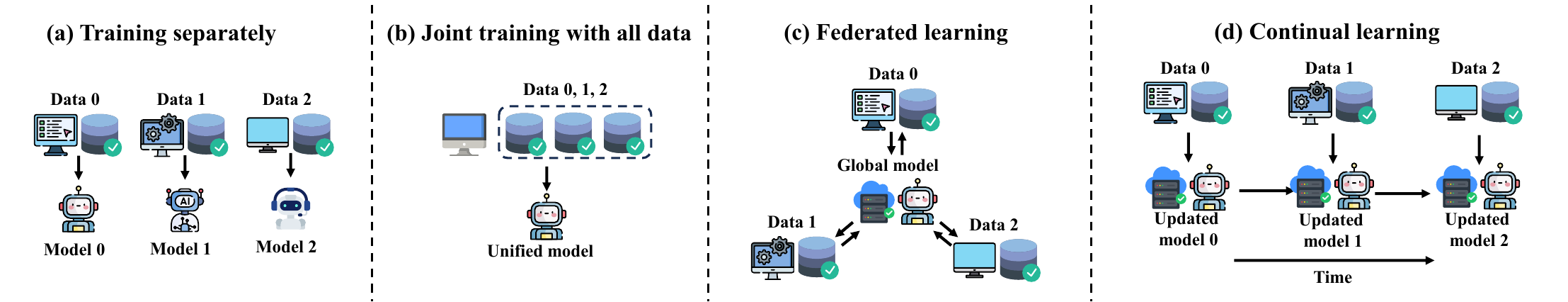}
  \caption{Method comparison with multiple mobility datasets.}
  \label{fig:different method}
\end{figure*}

To address these data barriers, recent studies have proposed different training strategies to leverage multiple datasets,
and Figure~\ref{fig:different method} compares different approaches. The most common strategy is to train models separately, as shown in Figure~\ref{fig:different method}(a). 
Recent efforts such as UniTraj~\cite{zhu2024unitraj}, TrajBert~\cite{si2023trajbert}, and TrajFM~\cite{lin2024trajfm} have explored joint training for unified representation and cross-city generalization (Figure~\ref{fig:different method}(b)), but these models remain tightly coupled with proprietary or limited-quality datasets.
TrajFM and TrajBert are typically pre-trained on restricted or private data. While UniTraj uses public data, it suffers from low semantic richness and high sparsity.
Federated learning~\cite{feng2020pmf,10598171LightTR} offers a potential solution for distributed mobility model training (Figure~\ref{fig:different method}(c)), but its reliance on frequent synchronization and communication poses challenges for scalability and practical deployment.
Consequently, these approaches fall short of meeting the diverse, dynamic, and multi-source demands of real-world mobility modeling, and cannot support an open ecosystem of shared models as seen in NLP and CV domains.

To truly usher in an era of shareable and sustainable human mobility science, we argue for a new collaborative paradigm.
This paradigm enables multiple data holders to jointly evolve and continually build a foundation model without sharing raw data, while preserving both privacy and generalization capability.
However, this vision poses several critical challenges:
Privacy preservation: How to achieve cross-institutional collaboration without any raw data exchange. (2) Cumulative knowledge retention: How to prevent ``catastrophic forgetting'' so that the model can accumulate knowledge from diverse urban systems over time. (3) Scientific heterogeneity: How to design an architecture that captures the unique characteristics of different urban structures (e.g., the grid layouts of New York vs. the radial patterns of Paris).

We propose MoveGCL, a scalable training framework for mobility foundation models based on generative continual learning. MoveGCL allows each data holder to evolve a shared model locally without exposing raw data, thereby ensuring full privacy preservation.
Specifically, MoveGCL starts from a pre-trained base model and employs a synthetic trajectory replay mechanism: instead of accessing historical data, each participant generates synthetic trajectories that approximate previously seen mobility patterns. This replay process preserves prior knowledge and mitigates catastrophic forgetting. Furthermore, knowledge distillation is applied during replay to reinforce the model’s ability to retain past capabilities while adapting to new data.
To handle the diversity of mobility data, MoveGCL adopts a Mixture-of-Experts (MoE) architecture equipped with a mobility pattern-aware expert routing mechanism. This design enables the model to dynamically select expert modules tailored to local mobility characteristics.
Together, these innovations make MoveGCL a practical and privacy-preserving solution for collaboratively building generalizable mobility foundation models across distributed, heterogeneous, and privacy-sensitive data sources.
In summary, our key contributions are as follows:
\begin{itemize}[leftmargin=*]
    \item We formalize the first privacy-preserving collaborative training paradigm for mobility foundation models, enabling the scientific community to build unified models across data silos.
    \item    We propose MoveGCL, a novel framework based on generative continual learning. Its core components—knowledge distillation, mobility-aware expert routing, and layer-wise progressive adaptation address catastrophic forgetting and urban heterogeneity, facilitating the long-term accumulation of mobility knowledge.
    \item  We demonstrate on six diverse global urban datasets that MoveGCL achieves performance comparable to centralized joint training, providing a practical pathway toward Open Mobility Science.
\end{itemize}

\section{Related Work}

Due to space limitations, we provide a concise review here and defer the extended discussion to Appendix~\ref{sec:related}.

\paragraph{\textbf{Mobility Data}}
Mobility data is commonly represented as aggregated flows or individual trajectories~\cite{barbosa2018human,yuan2025learning,zhang2025noise}. While flow data is relatively accessible and widely used in urban analytics~\cite{rong2024interdisciplinary,zhang2017deep}, trajectory data is often fragmented due to privacy constraints and institutional silos~\cite{kong2023mobility,yuan2025learning}. Existing real-world and open trajectory datasets are typically limited in city coverage, duration, or sampling density, and even recent global-scale datasets may suffer from inconsistent resolution and quality. Generative AI has enabled synthetic mobility datasets~\cite{zhu2023synmob,yabe2024yjmob100k,yuan2025worldmove}, but matching real trajectories remains challenging, especially in behavioral diversity and temporal continuity. As a result, many studies rely on restricted-access data and cannot release the trajectories they use~\cite{schlapfer2021universal}.

\paragraph{\textbf{Mobility Foundation Models}}
Recent work has begun to build foundation models for urban and mobility applications~\cite{yuan2024urbandit,chai2025mobiworld,han2025trajmoe,zhou2024urban,zhang2024urban,choudhury2024towards}. Most efforts focus on aggregated mobility signals and demonstrate strong cross-city transfer via unified spatio-temporal pre-training~\cite{li2024urbangpt,yuan2024uniflow,yuan2024unist,li2024opencity}. In contrast, foundation models for individual mobility remain less mature due to limited representative trajectory corpora and the complexity of human behavior~\cite{yuan2025learning,zhang2025noise,long2024universal}. LLM-based trajectory generation has also been explored~\cite{shao2024beyond,gong2024mobility,ju2025trajllm,feng2024agentmove,jiawei2024large}, but the modality gap suggests the need for native mobility foundation models that can later be aligned with language-based reasoning.

\paragraph{\textbf{Continual Learning}}
Continual (lifelong) learning aims to update models on sequential data without degrading previously learned capabilities~\cite{kim2022theoretical,wang2024comprehensive,yoon2021federated}. A key challenge is catastrophic forgetting~\cite{li2019learn,wickramasinghe2023continual}. Existing methods typically fall into regularization, replay, and parameter-isolation families, which trade off memory, computation, and stability when adapting to new data streams.

\section{Preliminaries}

\paragraph{\textbf{Data Format}}
In our setting, human mobility data is represented as sequences of spatiotemporal tokens, where each token corresponds to a visited location at a specific time. The spatial domain is typically discretized into a uniform grid (500m × 500m resolution), and the temporal domain is segmented into fixed-length intervals (30 minutes). Each individual trajectory can be formulated as a sequence ${z_1, z_2, \dots, z_T}$, where $z_t = (l_t, t_t)$ denotes the location and timestamp of a mobility event at time step $t$. These sequences capture rich behavioral patterns across time and space and form the foundation for model training.

\paragraph{\textbf{Model Training via Next-Token Prediction}}
Following the standard practice in language modeling, the training objective for mobility foundation models is formulated as a next‐location prediction task. Given a partial trajectory \(\{z_1, \dots, z_{t-1}\}\), the model aims to predict the next location \(l_t\), where \(l_t\) represents the spatial component of the upcoming step in the trajectory.
Formally, the training objective is defined as maximizing the log-likelihood of the observed sequence:
\begin{equation}
\mathcal{L} = \sum_{t=1}^{T} \log P(l_t \mid z_1, \dots, z_{t-1}; \theta),
\end{equation}
where $\theta$ denotes the model parameters. This objective allows the model to learn rich dependencies across spatial locations, temporal patterns, and contextual semantics, and serves as the core pretraining strategy for mobility foundation models.

\paragraph{\textbf{Training Pipeline for Mobility Foundation Model Development}}
We adopt a continual learning paradigm to train the mobility foundation model. To simulate learning on highly heterogeneous data, each round of continual learning introduces the dataset of a new city to the model.
Initially, the base model \(f_{\mathrm{base}}\) is trained using the base dataset \(\mathcal{D}_{\mathrm{base}}\). During continual learning, at the beginning of the \(n\)-th round, the model has already been trained on the dataset \(\mathcal{D}_{\mathrm{all,}n-1} = \mathcal{D}_{\mathrm{base}} \cup \mathcal{D}_{\mathrm{continual,}n-1}\), where \(\mathcal{D}_{\mathrm{continual},n-1} \;=\; \bigcup_{i=1}^{n-1} d_i\), and each \(d_i\) represents the dataset introduced in the \(i\)-th round. However, the historical dataset \(\mathcal{D}_{\mathrm{all,}n-1}\) is no longer accessible. The model update at round \(n\) is performed solely based on the current data \(d_n\) and a copy of the model from the previous round, denoted as \(f_{\mathrm{old,}n}\).The continual learning process can be formalized as the following optimization objective:
\begin{equation} 
\theta_{\mathrm{new,}n} = \arg\min_{\theta} \mathcal{L}\left(\theta;\, d_n;\, f_{\mathrm{old,}n}\right), \quad \text{s.t.} \quad \theta \gets \theta_{\mathrm{old,}n},
\end{equation}
where \(\theta\) denotes the model parameters, and \(\theta_{\mathrm{old},n}\) and \(\theta_{\mathrm{new},n}\) correspond to the parameters of \(f_{\mathrm{old},n}\) and \(f_{\mathrm{new},n}\), respectively. The loss function\(\mathcal{L}(\theta; d_n; f_{\mathrm{old,}n})\)
incorporates constraint terms derived from the previous model \(f_{\mathrm{old},n}\) to mitigate catastrophic forgetting. Rather than reinitializing from scratch, we optimize \(\theta_{\mathrm{new},n}\) starting from \(\theta_{\mathrm{old},n}\). Specifically, when \(n = 0\), we set \(f_{\mathrm{old},n} = f_{\mathrm{base}}\); for \(n > 0\), we have \(f_{\mathrm{old},n} = f_{\mathrm{new},\,n-1}\).

\begin{figure*}[t]
  \centering
  \includegraphics[width=\textwidth]{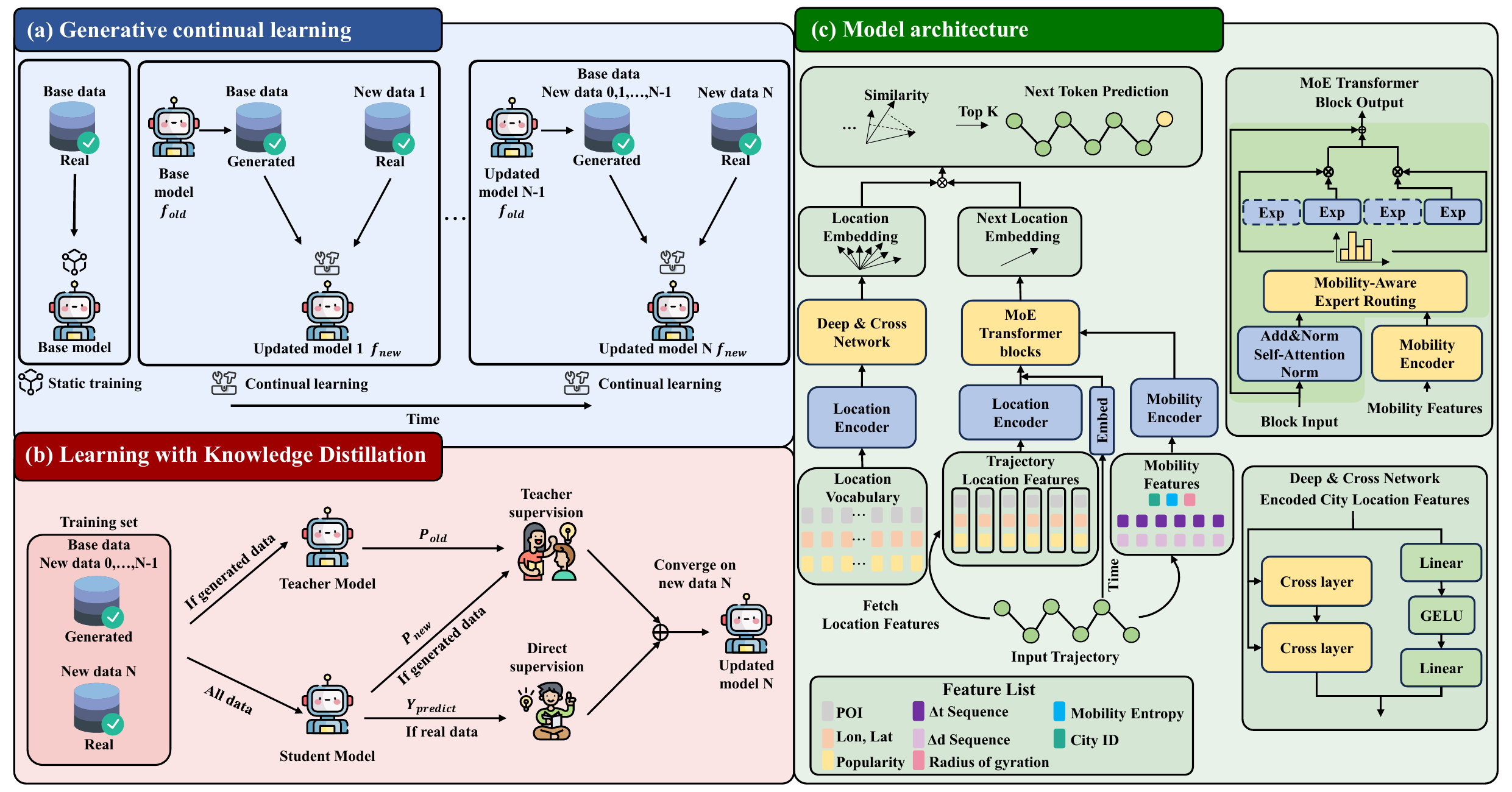}
  \caption{Overview of the MoveGCL framework: (a) the overall workflow; (b) the implementation of generative continual learning; (c) the model architecture.}
  \label{fig:Generative Continual Learning}
\end{figure*}
\section{MoveGCL}

In this section, we introduce the overall framework of MoveGCL, which is shown in Figure~\ref{fig:Generative Continual Learning}.

\subsection{Generative Continual Learning}\label{sec:GCL}


\paragraph{\textbf{Generative Replay with Teacher Model}}
To retain knowledge from previously visited cities without storing real-world mobility trajectory data, we design a generative replay strategy, as illustrated in Figure~\ref{fig:Generative Continual Learning}(a). At each stage, we keep a copy of the previously trained model $f_{\text{old}}$, referred to as the \textit{teacher model}. This teacher model represents the model trained on earlier mobility datasets. It remains frozen during subsequent learning and serves as a knowledge source to guide the student model $f_{\text{new}}$ when learning new cities.



To simulate past mobility behaviors, we employ \(f_{\text{old}}\) to generate synthetic trajectory data \(\tilde{x}_{\text{old}}^{c_i}\). First, we extract a trajectory
\begin{equation}
x_{\text{new}} = \bigl[(l'_0,\,t'_0),\,(l'_1,\,t'_1),\,\ldots,\,(l'_{L},\,t'_{L})\bigr],
\end{equation}
from the new dataset, where \(L\) denotes its length and \(\{(l'_i,\,t'_i)\}_{i=0}^{L}\) are the location–time pairs. Next, for a specific previously observed city \(c_i\), we sample an initial location from the empirical distribution of actual locations in city \(c_i\) conditioned on length \(L\):
$l_0 \;\sim\; \rho_{\text{loc}\mid L}^{c_i}\,$,
where \(\rho_{\text{loc}\mid L}^{c_i}\) denotes the empirical distribution of initial  locations in previously observed city \(c_i\) given a trajectory length \(L\). We then replace \(l'_0\) with the sampled \(l_0\) and generate the pseudo old‐city trajectory
\[
\tilde{x}_{\text{old}}^{c_i} \;=\; \bigl[(l_0,\,t'_0),\,(l_1,\,t'_1),\,\ldots,\,(l_{L},\,t'_{L})\bigr],
\]
by drawing subsequent locations autoregressively:
\begin{equation}
(l_1,\,l_2,\,\ldots,\,l_{L}) \;\sim\; f_{\text{old}}\Bigl(\cdot \,\bigm|\;l_0,\;\{t'_i\}_{i=0}^{L}\Bigr),
\end{equation}
where each \(t'_i\) is taken from the time distribution of the extracted new‐data trajectory.  

These pseudo‐trajectories reflect the mobility patterns learned in earlier stages. We combine all pseudo‐trajectories with the real data from the current city, to construct the full training set:
\begin{equation}
\mathcal{D}_{\text{train}} \;=\; \bigcup_{i=1}^{N} \alpha\,\tilde{\mathcal{X}}_{\text{old}}^{c_i} \;\cup\; \mathcal{X}_{\text{new}}\,,
\end{equation}
where \(\{c_1, c_2, \ldots, c_N\}\) denotes the \(N\) previously observed cities, \(\tilde{\mathcal{X}}_{\text{old}}^{c_i}\) denotes the set of pseudo‐trajectories generated for city \(c_i\), and \(\mathcal{X}_{\text{new}}\) is the set of real trajectories from the current city. Coefficient \(\alpha>0\) specifies ratio between the number of pseudo‐trajectories in each previous city and the number of real trajectories in \(\mathcal{X}_{\text{new}}\).

This allows the student model to learn current-city behaviors while retaining knowledge of previously learned cities.

\paragraph{\textbf{Distilling Knowledge to Preserve Mobility Patterns}}
To further strengthen the model's ability to preserve prior knowledge, we introduce a knowledge distillation loss that transfers behavioral patterns from the teacher model to the student model, as depicted in Figure~\ref{fig:Generative Continual Learning}(b). For each generated pseudo-trajectory $\tilde{x}_{\text{old}}$, we extract the predicted mobility distributions from both models:
\begin{equation}
P_{\text{old}}(\cdot \mid \tilde{x}_{\text{old}}) = f_{\text{old}}(\tilde{x}_{\text{old}}), \quad 
P_{\text{new}}(\cdot \mid \tilde{x}_{\text{old}}) = f_{\text{new}}(\tilde{x}_{\text{old}}).
\end{equation}
We minimize the Kullback–Leibler (KL) divergence between the teacher’s and student’s predicted distributions:
\begin{equation}
\mathcal{L}_{\text{KD}} = \mathbb{E}_{\tilde{x}_{\text{old}} \sim f_{\text{old}}} \left[ 
\mathrm{KL} \left( P_{\text{old}}(\cdot \mid \tilde{x}_{\text{old}}) \,\|\, P_{\text{new}}(\cdot \mid \tilde{x}_{\text{old}}) \right) 
\right].
\end{equation}

For the new data \(x_{\mathrm{new}}\), we compute the task loss as the cross‐entropy between the model’s predicted distribution and the true labels, referred to as
\(\mathcal{L}_{\text{cross‐entropy}}\).
The total training objective of the student model is a weighted sum of the task loss on new-city data and the distillation loss:
\begin{equation}
\mathcal{L}_{\text{total}} = \mathcal{L}_{\text{cross‐entropy}} + \lambda \cdot \mathcal{L}_{\text{KD}},
\end{equation}
where $\lambda$ is a hyperparameter that balances learning new knowledge and retaining previously acquired behaviors.

\subsection{Model Architecture}\label{sec:model}

To enable scalable and adaptive learning across heterogeneous urban environments, our model is designed with a modular architecture that integrates flexible location encoders/decoders, a Mixture-of-Experts (MoE) Transformer backbone, and a mobility-aware expert routing mechanism, as shown in Figure~\ref{fig:Generative Continual Learning}(c). This design ensures the model's scalability and adaptability in multi-city continual learning scenarios.

\paragraph{\textbf{Unified Location Encoder}}
Conventional location representations often rely on discrete location IDs, which are inherently city-specific and hinder cross-city generalization. To overcome this limitation, we design a continuous location representation that embeds each location into a shared latent space, capturing transferable semantic and spatial properties across cities. This unified representation facilitates knowledge sharing and supports incremental learning across heterogeneous urban environments.
Concretely, each location $l \in \mathcal{L}$ is represented by a feature vector $\mathbf{z}_l \in \mathbb{R}^d$ constructed from three key components:
$\mathbf{z}_l = \phi_{\text{POI}}(l) \oplus \phi_{\text{lat-lon}}(l) \oplus \phi_{\text{hot}}(l),$
\noindent where $\phi_{\text{POI}}(l) \in \mathbb{R}^{d_1}$ denotes the Point-of-Interest (POI) embedding, capturing semantic land-use attributes such as residential, commercial, educational, or recreational functions, often indicative of mobility intent and purpose; $\phi_{\text{hot}}(l) \in \mathbb{R}^{d_2}$ is the mobility heat embedding, derived from public available OD flows at each location, which reflects the functional centrality of that location; $\phi_{\text{lat-lon}}(l) \in \mathbb{R}^{d_3}$ is the normalized latitude-longitude embedding, representing the relative spatial position of the location within the city boundary.

The overall location representation $\mathbf{z}_l$ is obtained via concatenation ($\oplus$) of these features, and is further transformed by a shared multi-layer perceptron (MLP). This design is sufficient and generalizable because it captures three complementary views of spatial semantics: (1) semantic functions via POI types, (2) actual mobility signals via visitation popularity, and (3) captures where the location sits in the urban layout. Together, they provide a compact yet expressive embedding that generalizes well across different cities with varied spatial structures.

\begin{figure*}[t!]
  \centering
  \includegraphics[width=0.95\linewidth]{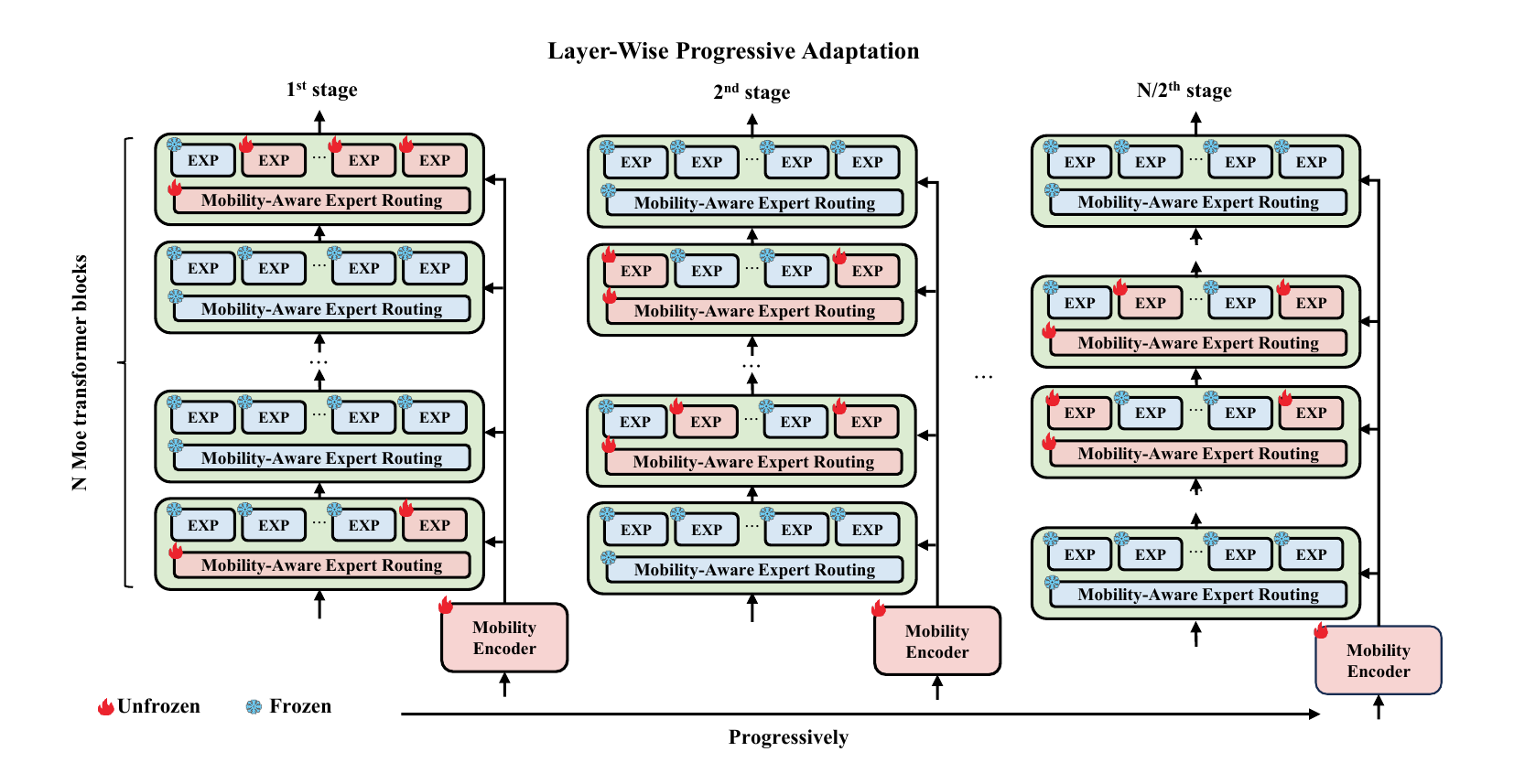}
  \vspace{-3mm}
  \caption{Illustration of the layer-wise progressive adaptation}
  \label{fig:Layer-wise Progressive Adaptation}
\end{figure*}

\paragraph{\textbf{Mixture-of-Experts Transformer}}
The Mixture-of-Experts (MoE) architecture comprises a router network and multiple expert networks, serving as a replacement for the Feed-Forward Network (FFN) within the Transformer~\cite{masoudnia2014mixture}. The output of the MoE layer, $F_{\mathrm{MoE}}(x)$, is the weighted sum of the selected expert outputs, where the weights are given by the router network’s output:
\begin{equation}
    F_{\mathrm{MoE}}(x) \;=\; \sum_{i=1}^{k} R_i(x)\,\cdot\,E_i(x).
\end{equation}
Here, $x$ denotes the input to the MoE module, $k$ is the number of selected experts, $R_i(x)$ is the output of the router network for expert $i$ (detailed in Section~\ref{part:Mobility-Aware Expert Routing}), $E_i(x)$ is the output of expert $i$.

MoveGCL is built upon  Mixture-of-Experts (MoE) Transformer blocks, in which each expert module is responsible for capturing specific mobility patterns. During continual learning, we introduce new experts to accommodate knowledge from new cities, and design layer-wise progressive adaptation training strategy (detail in Section~\ref{sec:training}). This partial parameter update strategy injects new knowledge without overwriting existing capabilities, thus alleviating catastrophic forgetting. The modularity of the MoE block also supports elastic model expansion as more cities are introduced.

\paragraph{\textbf{Mobility-Aware Expert Routing}}\label{part:Mobility-Aware Expert Routing}
For each input trajectory \(x\), we extract a set of mobility behavior descriptive features and encode them into a mobility feature descriptor vector \(\mathbf{z}_m \in \mathbb{R}^d\). This feature set comprises: the jump distance \(d_{\mathrm{jump}}\) (distance between the current point and the previous point in the trajectory); the waiting time \(t_{\mathrm{wait}}\) (time difference between arrivals at the current and previous points in the trajectory); the quantized radius of gyration \(r_{\mathrm{gyr}}\); the quantized location entropy \(H_{\mathrm{loc}}\); and the city identifier \(\mathrm{ID}_{\mathrm{city}}\). These features are embedded via their respective encoders, where \(d_{\mathrm{jump}}\) and \(t_{\mathrm{wait}}\) are processed by a Transformer-based continuous feature encoder, and \(r_{\mathrm{gyr}}\), \(H_{\mathrm{loc}}\), and \(\mathrm{ID}_{\mathrm{city}}\) are handled by discrete embedding modules. Finally, the five feature embeddings are concatenated to form the mobility behavior vector:
\begin{equation}\label{eq:mobility_feature}
\mathbf{z}_m \;=\;
\begin{bmatrix}
    \phi_{d_{\mathrm{jump}}}\bigl(d_{\mathrm{jump}}(x)\bigr),\,
    \phi_{t_{\mathrm{wait}}}\bigl(t_{\mathrm{wait}}(x)\bigr),\,
    \phi_{r_{\mathrm{gyr}}}\bigl(r_{\mathrm{gyr}}(x)\bigr), \\[6pt]
    \phi_{H_{\mathrm{loc}}}\bigl(H_{\mathrm{loc}}(x)\bigr),\,
    \phi_{\mathrm{ID}_{\mathrm{city}}}\bigl(\mathrm{ID}_{\mathrm{city}}\bigr)
\end{bmatrix}.
\end{equation}
Here, \(\phi_{d_{\mathrm{jump}}}\) and \(\phi_{t_{\mathrm{wait}}}\) denote the Transformer encoders for continuous mobility features, while \(\phi_{r_{\mathrm{gyr}}}\), \(\phi_{H_{\mathrm{loc}}}\), and \(\phi_{\mathrm{ID}_{\mathrm{city}}}\) represent the embedding encoders for discrete features.

Within each layer of the MoE blocks, we introduce a routing network based on a linear transformation to compute the routing weights for each expert, leveraging both the mobility descriptor feature vector \(\mathbf{z}_m\) and the output of the self-attention submodule at that layer. The routing weights for layer \(i\) are computed as:
\begin{equation}\label{eq:routing_weight}
    R_i(x) = \mathsf{softmax}\Bigl(\mathsf{TopK}\bigl(W_{i,r}\,(\mathbf{z}_m \oplus X_i(x))\;+\;b_i\bigr)\Bigr),
\end{equation}
where \(\mathbf{z}_m\) is the mobility feature descriptor vector defined above; \(X_i(x)\) denotes the output of the self-attention submodule; \(W_{i,r}\) is a learnable projection matrice; \(b_i\) is the bias term; and \(\mathsf{TopK}(\cdot)\) retains only the top \(K\) values (corresponding to the \(K\) highest-scoring experts), setting the remaining expert scores to \(-\infty\) so that their weights after the softmax operation are effectively zero.

This routing strategy achieves two goals: (1) it promotes functional specialization by directing similar motion patterns to consistent expert subsets; (2) it enables the model to discover and transfer shared mobility structures across cities, thus enhancing generalization in multi-city scenarios. Additionally, this mobility-aware routing provides a structured inductive bias that accelerates model adaptation during incremental learning, allowing new experts to specialize rapidly with minimal interference to retained knowledge.

\paragraph{\textbf{Similarity–Based Decoder}}
The next-location prediction is performed by computing the similarity between the final output of the Mixture-of-Experts (MoE) Transformer blocks and the representation vectors of all candidate locations in the city.  
These location representations are generated using a Deep \& Cross Network (DCN)~\cite{wang2017deep}, which consists of both a Cross layer and a Deep layer to capture feature interactions and nonlinear transformations.

The Cross network captures inter-location correlations within a city by applying element-wise interactions over location embeddings \(E_l\), producing:
\begin{equation} 
E_{\mathrm{cross}} = \sum_{i=1}^{d} E_{i} \odot W_{i} E_{l} + b_{i},
\end{equation}
where \(\odot\) denotes element-wise multiplication, and \(W_i\), \(b_i\) are learnable parameters.
The Deep layer refines the same \(E_l\) using a two-layer MLP:
\begin{equation} 
E_{\mathrm{deep}} = \mathrm{GELU}\left(\left(W_1 E_l + b_1\right) W_2 + b_2\right),
\end{equation}
with weights \(W_1, W_2\), biases \(b_1, b_2\), and GELU activation.
The DCN output is the concatenation of both branches:
\begin{equation}
E_{\mathrm{DCN}} = E_{\mathrm{cross}} \oplus E_{\mathrm{deep}}.
\end{equation}

For next-location prediction, the user's historical trajectory is encoded by MoE transformer blocks to produce a prediction vector \(P\); separately, the Deep \& Cross Network (DCN) encodes every candidate location in the city to produce location representations \(E_{\mathrm{DCN}}\). The similarity score is computed as:


\begin{equation}
\mathrm{Score}_{\mathrm{similarity}} = \sum_{i=1}^{d} P \cdot E_{\mathrm{DCN},i},
\end{equation}
where \(d\) is the number of locations. Higher similarity scores indicate higher probabilities of being the next location.
This similarity-based decoding strategy ensures scalability, as it decouples the prediction process from a fixed output space. Instead of classifying over a static set of locations, the model performs representation-level matching, allowing it to generalize across cities with different spatial layouts and dynamically varying numbers of candidate locations.

\begin{table*}[t!]
\caption{Performance comparison between MoveGCL and baseline methods across different cities and Acc@k metrics.}
\label{tbl:main_result}
\vspace{-3mm}
  \centering
\resizebox{\textwidth}{!}{
\begin{tabular}{c cc cc cc cc cc cc}
\hline
\textbf{Model} & \multicolumn{2}{c}{\textbf{Atlanta}} & \multicolumn{2}{c}{\textbf{Chicago}} & \multicolumn{2}{c}{\textbf{Los Angeles}} & \multicolumn{2}{c}{\textbf{New York}} & \multicolumn{2}{c}{\textbf{Seattle}} & \multicolumn{2}{c}{\textbf{Washington D.C}} \\
\cmidrule(lr){2-3} \cmidrule(lr){4-5} \cmidrule(lr){6-7} \cmidrule(lr){8-9} \cmidrule(lr){10-11} \cmidrule(lr){12-13}
 & Acc@1 & Acc@3 & Acc@1 & Acc@3 & Acc@1 & Acc@3 & Acc@1 & Acc@3 & Acc@1 & Acc@3 & Acc@1 & Acc@3\\
\hline
Markov   &0.183 & 0.325 & 0.146 & 0.260 & 0.103 & 0.201 & 0.115 & 0.275 & 0.202 & 0.318 & 0.162 & 0.347 \\
LSTM     &0.231 & 0.373 & 0.194 & 0.334 & 0.131 & 0.275 & 0.169 & 0.312 & 0.259 & 0.420 & 0224 & 0.383\\
Transformer &0.210 & 0.353 & 0.175 & 0.300 & 0.124 & 0.268 & 0.156 & 0.318 & 0.235 & 0.397 & 0.192 & 0.356\\
DeepMove &0.242 & 0.393 & 0.203 & 0.344 & 0.147 & 0.274 & 0.177 & 0.329 & 0.278 & 0.444 & 0.247 & 0.408\\
TrajBert &0.214 & 0.370 & 0.183 & 0.316 & 0.146 & 0.277 & 0.159 & 0.310 & 0.234 & 0.402 & 0.207 & 0.367\\
CLET &0.263 & 0.422 & 0.200 & 0.341 & 0.138 & 0.275 & 0.156 & 0.313 & 0.289 & 0.454 & 0.232 & 0.390\\
\cmidrule(lr){1-13}
PMF &0.248&0.381&0.151&0.249&0.112&0.190&0.150&0.257&0.269&0.418&0.217&0.349\\
LightTR &0.269&0.402&0.168&0.269&0.130&0.216&0.168&0.281&0.296&0.444&0.242&0.376\\
\cmidrule(lr){1-13}
\makecell{MoveGCL\\(FullTune)}      &0.188 & 0.304 & 0.125 & 0.206 & 0.064 & 0.114 & 0.208 & 0.329 & 0.199 & 0.318 & 0.147 & 0.257 \\
\makecell{MoveGCL\\(ExpertTune)}      &0.192 & 0.310 & 0.132 & 0.215 & 0.062 & 0.108 & 0.207 & 0.327 & 0.195 & 0.322 & 0.147 & 0.259 \\
\cmidrule(lr){1-13}
\makecell{MoveGCL\\(WSC$\to$A$\to$L$\to$N)} &0.282 & 0.421 & 0.197 & 0.306 & 0.157 & 0.254 & 0.206 & 0.328 & 0.324 & 0.478 & 0.273 & 0.413 \\
\hline
\end{tabular}
}
\end{table*}

\subsection{Layer-Wise Progressive Adaptation}\label{sec:training}
To ensure a balance between previously and newly learned knowledge, MoveGCL employs a layer-wise progressive adaptation strategy, where model parameters are updated in stages, as illustrated in Figure \ref{fig:Layer-wise Progressive Adaptation}. For a model composed of $N$ layers of MoE transformer blocks, the total number of training epochs $E$ is evenly divided into $N/2$ stages, with each stage lasting $\frac{\text{E}}{N/2}$ epochs. At each stage, a pair of symmetrically positioned MoE transformer blocks—one near the input side and the other near the output side—are unfrozen for fine-tuning, while the remaining layers remain frozen. The specific process is as follows:

\begin{itemize}[leftmargin=*]
    \item In Stage 1, the outermost layers (closest to the input and output) are unfrozen.
    \item In Stage 2, the second closest layers to the input and output are unfrozen.
    \item ...
    \item In Stage N/2, the two central layers of the model are unfrozen.
\end{itemize}

Within each stage, only a subset of parameters in the unfrozen layers is updated—the routing modules, newly added experts and previously trained experts that were not frequently activated in the prior generative continual learning phase. To facilitate adaptation to the mobility features across different datasets, parameters of the mobility feature encoder are updated continuously during all stages. Furthermore, to prevent large parameter shifts during the initial stage of parameter updating, all previously trained experts in the input-side MoE transformer layer are kept frozen during Stage 1.

\section{Results}

\subsection{Experimental Settings}

\textit{\textbf{Datasets.}} We utilize human mobility datasets from multiple cities to evaluate the performance of MoveGCL. Specifically, the datasets cover over eight hundred thousand users and feature a relatively high sampling rate compared to currently available public datasets. Detailed statistics and descriptions are provided in Appendix Table~\ref{tab:Basic statistics of mobility data}.
For each city, we randomly sample 120,000 trajectories for training, 40,000 for validation, and 40,000 for testing.



\textbf{\textit{Baselines and Parameter Settings.}} We compare our method with a diverse set of baselines, including traditional mobility models, federated learning-based approaches, and joint learning models with privacy-preserving mechanisms. Appendix~\ref{sec:baseline} provides detailed descriptions of the baselines, and Appendix~\ref{sec:para} summarizes the parameter settings.

\subsection{Overall Performance}

Table~\ref{tbl:main_result} presents the performance of MoveGCL compared to state-of-the-art baseline methods. 

\begin{itemize}[leftmargin=*]
    \item \textbf{MoveGCL consistently outperforms traditional deep learning models trained independently on each dataset, demonstrating strong cross-city scalability.} On average, it achieves an 8\% improvement in Acc@1, highlighting its ability to generalize across diverse urban environments. This result validates the promise of mobility foundation models, which unify knowledge across cities and reduce redundancy. In contrast, training separate models for each city not only increases computational and deployment costs, but also fails to leverage shared mobility patterns across domains.
    \item  \textbf{MoveGCL surpasses privacy-preserving federated learning approaches in both accuracy and stability.} Compared to domain-specific baselines such as PVM~\cite{feng2020pmf} and LightTR~\cite{10598171LightTR}, MoveGCL achieves significantly higher performance. This advantage stems from its unified generative continual learning framework, which maintains global generalization without suffering from the synchronization overhead and convergence instability inherent in federated setups. This further supports its practicality in real-world multi-party mobility modeling scenarios.
    \item \textbf{MoveGCL effectively balances adaptation to new data while retaining knowledge from previously seen data.} We benchmark against two continual learning strategies—FullTune and ExpertTune—which either fine-tune or expand the model on new datasets. While these methods partially adapt to new cities, they suffer from severe performance degradation on previously seen data, indicating catastrophic forgetting. In contrast, MoveGCL preserves prior knowledge and achieves higher performance on both old and new datasets, demonstrating its ability to support continuous model evolution without sacrificing stability.
\end{itemize}

\subsection{Order Invariance in Continual Learning}
To assess the robustness of MoveGCL in real-world deployment scenarios, we investigate its sensitivity to the order in which data from different cities is introduced during continual learning.  As shown in Appendix Table~\ref{tbl:order}, the performance of MoveGCL remains remarkably consistent, with the vast majority of metrics showing deviations of less than 5\% across original and reversed sequences. This empirical finding confirms the order-invariant learning behavior of our model, demonstrating that MoveGCL can integrate new city data without disrupting previously acquired knowledge, even when the order of exposure varies significantly.  We refer the reader to the detailed analysis in Appendix~\ref{sec:order}.


\subsection{Privacy Evaluation}

As MoveGCL is built on a generative continual learning framework that does not retain raw data from previous cities, a key question is whether the synthetic data used for replay may inadvertently leak private information from the original training data. To rigorously evaluate the privacy-preserving properties of our approach, following the methodology in~\cite{yuan2025learning,yuan2024needdyn}, we conduct a comprehensive analysis from three complementary perspectives:

\begin{figure}[t!]
  \centering
  \vspace{-3mm}
  \includegraphics[width=\linewidth]{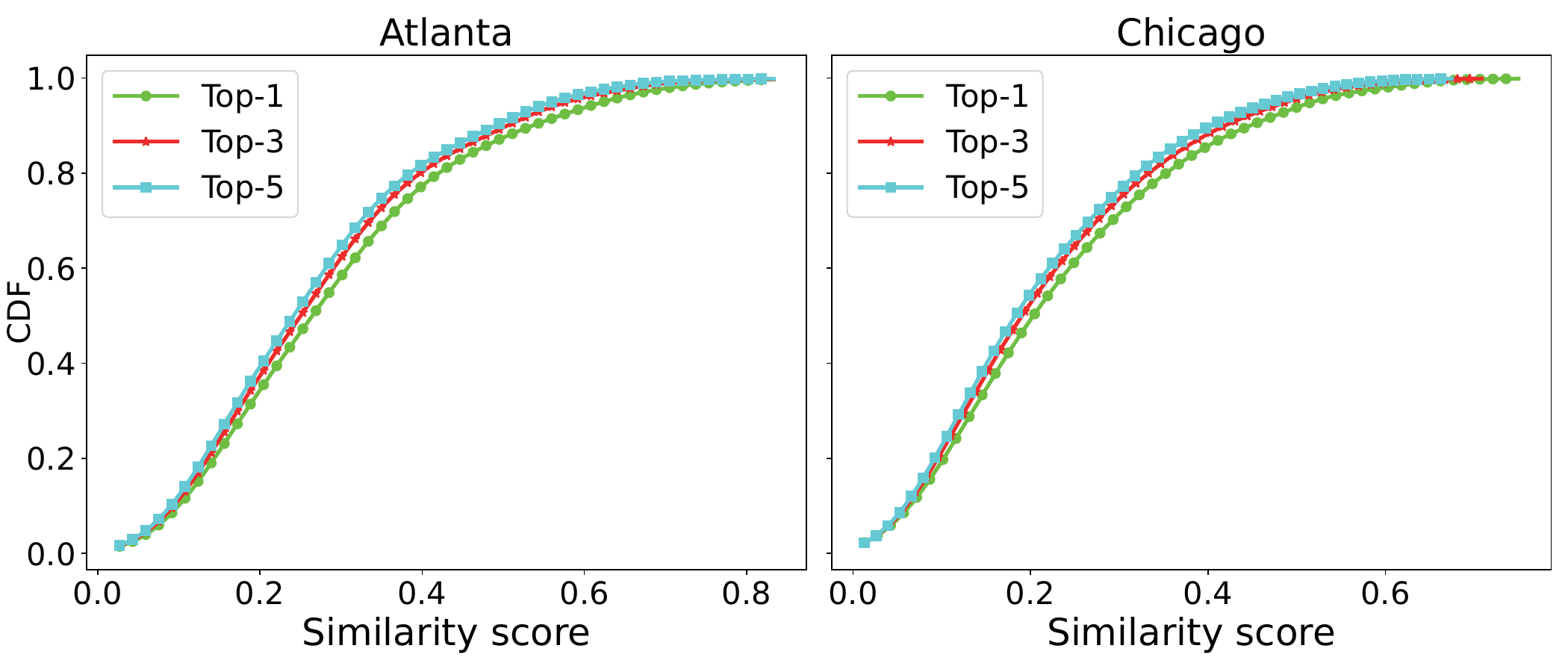}
  \vspace{-4mm}
  \caption{Similarity score distribution in uniqueness testing.}
  \label{fig:Uniqueness Testing}
\end{figure}

\begin{figure}[t!]
  \centering
  \vspace{-3mm}
  \includegraphics[width=0.9\linewidth]{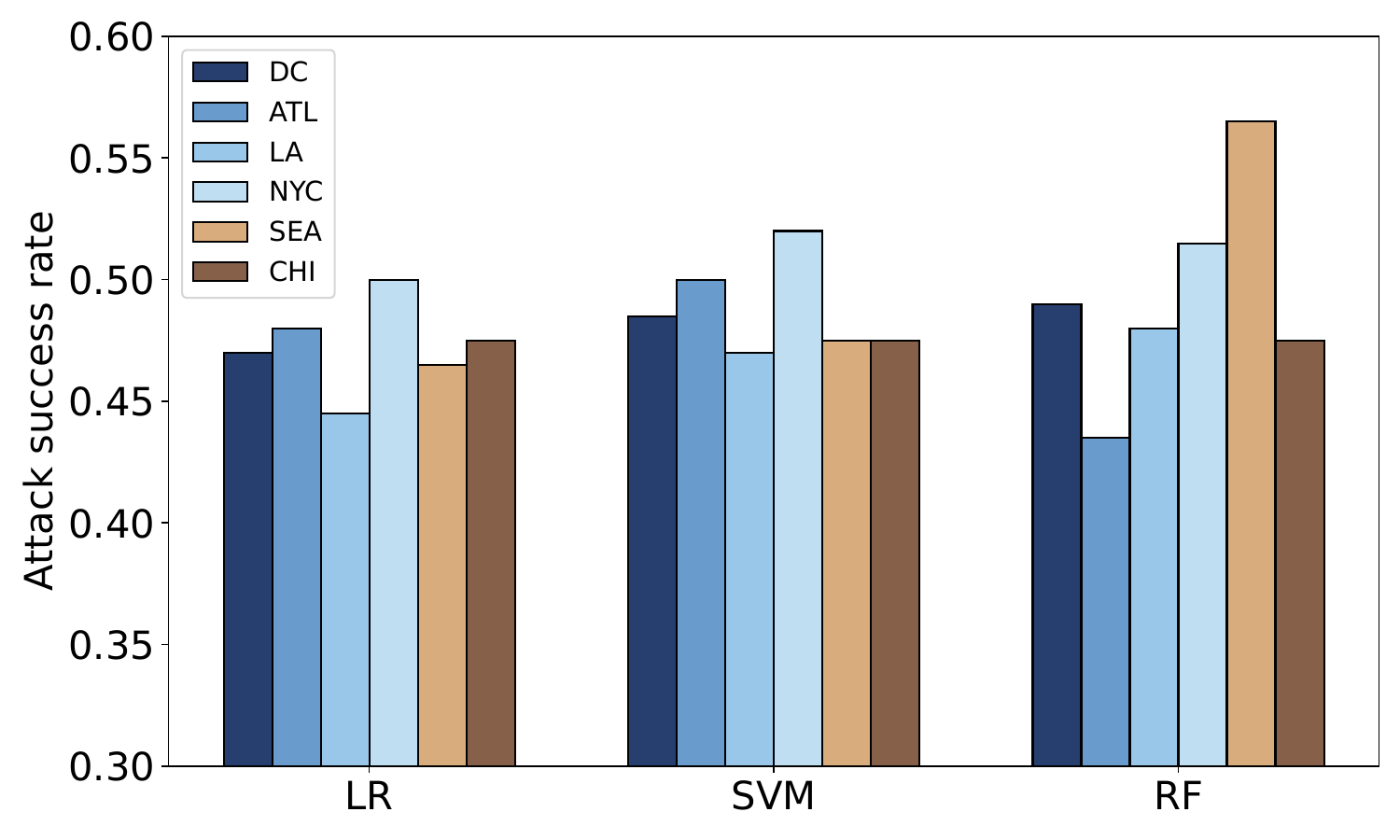}
  \vspace{-2mm}
  \caption{Success rate in membership inference attack}
  \label{fig:Membership Inference Attack}
  \vspace{-4mm}
\end{figure}

\begin{itemize}[leftmargin=*]
     \item \textbf{Uniqueness Testing}~\cite{demontjoye2013unique,xu2017trajectory}: To evaluate the degree of similarity between the generated data and the real data.
    \item \textbf{Membership Inference Attack}~\cite{lin2020using,shokri2017membership}: Given a trained model and a set of samples, it assesses whether an classifier can accurately determine which samples were included in the model’s training set based on the model’s outputs.
     \item \textbf{Differential Privacy}~\cite{abadi2016deep,andrew2019tensorflow}: To ensure that the model does not depend on a small subset of training examples, we remove a minimal set of training samples and evaluate whether the distribution of model outputs undergoes an obvious change.
\end{itemize}

Due to space limitations, we summarize the key conclusions here, and provide detailed experimental settings for the privacy evaluation in Appendix~\ref{sec:set_privacy}.




\textit{\textbf{Uniqueness Testing.}}
Figure~\ref{fig:Uniqueness Testing} presents the cumulative distribution of similarity scores. As shown in the figure, over 95\% of the generated trajectories do not have any corresponding real trajectory with a similarity score higher than 50\%. This indicates that the model’s outputs are based on the knowledge it has acquired rather than directly copying trajectories from the training set.


\textit{\textbf{Membership Inference Attack.}}
Figure~\ref{fig:Membership Inference Attack} shows the attack results. As observed, the success rates across all datasets are approximately 50\%, indicating that the classifier can hardly determine whether a trajectory was part of the training data or not based on the generated sample. These results indicate that our model is not easily susceptible to membership inference attacks.

\begin{table}[t!]
  \caption{Differential Privacy statistics by city.}
  \label{tab:Differential Privacy statistics by city}
  \vspace{-3mm}
  \centering
  \begin{tabular}{l|c|c|c}
    \hline
    $\boldsymbol{\epsilon}$ & Mean & Median & 75th Percentile \\
    \hline
    Atlanta        &2.671        &0.706        &2.212        \\
    Chicago        &2.919        &0.752        &2.504        \\
    Los Angeles    &2.988        &0.593        &2.572        \\
    New York       &3.394        &0.713        &2.001        \\
    Seattle        &2.934       &0.655        &1.870       \\
    Washington D.C &3.037        &0.600        &1.787        \\
    \hline
  \end{tabular}
  \vspace{-2mm}
\end{table}

\begin{figure*}[t]
  \centering
  \includegraphics[width=0.9\textwidth]{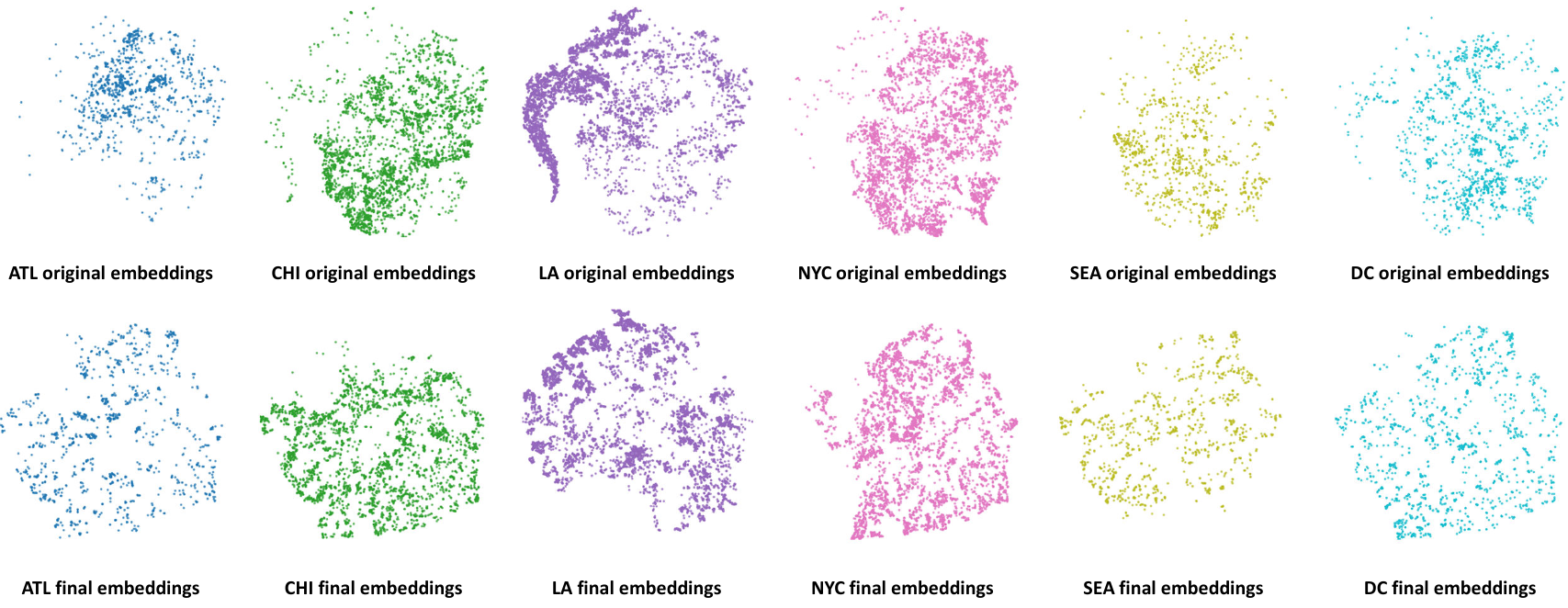}
  \caption{City location embeddings before (original) and after (final) DCN.}
  \label{fig:Location Encoding}
\end{figure*}

\textit{\textbf{Differential Privacy.}}
As shown in Table~\ref{tab:Differential Privacy statistics by city}, without applying any additional privacy-preserving mechanisms, 
MoveGCL naturally achieves a privacy budget of $\varepsilon \approx 1\text{--}2$ for 75\% of randomly sampled trajectories.
This level is generally considered an acceptable operating point for generative models~\cite{lin2020using}; 
for example, Apple adopts a privacy budget of $\varepsilon = 4.0$
\footnote{\url{https://www.apple.com/privacy/docs/Differential_Privacy_Overview.pdf}}.

\subsection{Ablation Studies}
In this section, we conduct two sets of incremental ablation studies based on MoveGCL (WSC→A→L→N). The first set focuses on the input features of the Mobility-Aware Expert Routing module. We selectively remove or adjust different dimensions of the location feature to evaluate the contribution and necessity of each type of input in guiding expert routing. The second set targets the incremental learning mechanism itself. We remove the knowledge distillation strategy designed to mitigate catastrophic forgetting in GCL, and instead train the model using only the conventional cross-entropy loss. This setup allows us to assess the effectiveness of knowledge distillation in preserving previously learned knowledge.

As shown in Appendix Figure~\ref{fig:Ablation_Studies}, removing any input feature from the Mobility-Aware Expert Routing module leads to a noticeable performance drop. Similarly, disabling the knowledge distillation strategy in GCL also results in a significant decline in model performance. These findings highlight the critical role of each input feature in expert selection, as well as the importance of knowledge distillation in ensuring model stability during continual learning.

\subsection{In-Depth Analysis}

To better understand why MoveGCL is capable of unifying diverse mobility datasets and effectively handling substantial inter-city heterogeneity, we conduct an in-depth analysis of the location embedding layer to examine whether MoveGCL can learn shared spatial representations across cities.
To this end, we extract the location embeddings for each city at two stages: (1) after the initial encoder, and (2) after the Deep \& Cross Network (DCN). By comparing these two sets of embeddings, we  evaluate the role of DCN in aligning spatial semantics across heterogeneous urban environments. 

As shown in Figure~\ref{fig:Location Encoding}, DCN aligns location embedding distributions in different cities much more closely than the original encoder output. This indicates that DCN successfully captures shared location-feature patterns across urban areas, thereby boosting the model’s ability to generalize in cross‐city settings. Moreover, within each city, the DCN‐processed embeddings become less densely clustered than their original counterparts, indicating a marked increase in separability among individual locations and further enhancing the model’s capacity to encode location semantics.

\subsection{Impact of Replay Volume}


In generative continual learning, synthetic data replay serves as a key mechanism for preserving previously acquired knowledge without accessing raw data. A critical hyperparameter in this process is the volume of generated data used during training on new cities. While too little replay data may result in catastrophic forgetting, excessive generation increases computational costs and may introduce noise or redundancy. Understanding this trade-off is essential for building scalable and efficient mobility foundation models.
To explore this, we vary the amount of generated data and evaluate its impact on both knowledge retention (for previously seen cities) and adaptation to new cities.

As shown in Appendix Figure~\ref{fig:data_scale}, the performance on the base cities (WSC) consistently improves with more generated data, indicating that replay volume directly affects the  ability to retain past knowledge. In contrast, performance on the newly introduced cities (A, L, N) remains largely stable regardless of replay volume, with no consistent trend of improvement or degradation.
These results suggest that while synthetic replay is crucial for mitigating forgetting, it has limited effect on new knowledge acquisition. Thus, allocating a moderate amount of generated data offers a practical balance—sufficient to preserve prior knowledge without incurring unnecessary overhead—supporting the long-term scalability.

\section{Limitations and Ethical Considerations}

While MoveGCL provides a robust framework for bridging data silos in mobility science, it faces several limitations and ethical challenges. First, the scientific generalizability of the model is inherently tied to the diversity of participating institutions; if the collaborative training sequence is skewed toward megacities in developed regions, it may fail to accurately represent the unique mobility dynamics of the Global South or rural areas, potentially leading to inequitable urban planning recommendations. From an ethical standpoint, although generative replay provides a significant privacy buffer, it still necessitates future research into integrating formal Differential Privacy (DP) guarantees. Furthermore, high-fidelity mobility foundation models present dual-use concerns; while they are intended to advance sustainability goals such as carbon footprint reduction, strict governance is required to prevent their misuse for unauthorized surveillance or restrictive movement policies. 

\section{Conclusion}

In this work, we present  a scalable and privacy-preserving framework for training mobility foundation models via generative continual learning. By enabling decentralized model evolution without sharing raw data, it addresses key challenges in real-world human mobility modeling, including data silos, privacy constraints, and heterogeneous mobility distributions.
MoveGCL represents a significant step toward realizing mobility foundation models  by offering a practical and generalizable framework that facilitates collaborative learning across cities and institutions. It paves the way for long-term, privacy-safe, and adaptive modeling of human movement, with broad implications for urban planning, transportation optimization, and evidence-based policy making. The development of more large-scale, semantically rich, and geographically diverse mobility datasets will be crucial for further improving the model’s generalization and robustness. We encourage the broader research community and data-holding institutions to join this collaborative effort, contributing to the creation of open, inclusive, and powerful spatiotemporal foundation models for the mobility domain.


\balance
\bibliographystyle{ACM-Reference-Format}
\bibliography{10-reference}

\clearpage
\appendix

\section{Experimental Settings}

\subsection{Dataset}

\begin{table}[H]
\caption{Basic statistics of mobility data.}
  \label{tab:Basic statistics of mobility data}
  \centering
  \begin{tabular}{cccc}
    \hline
    City & User & Trajectory & Location \\
    \hline
    Atlanta        &114941        &2348218        &1175        \\
    Chicago        &148000        &8051522        &4166        \\
    Los Angeles    &161544        &16844127        &6198        \\
    New York       &170321        &15766369        &4988        \\
    Seattle        &88569        &3362353        &1046       \\
    Washington D.C &134442        &11024181        &1361        \\
    \hline
  \end{tabular}
\end{table}

\subsection{Baselines}\label{sec:baseline}
\begin{itemize}[leftmargin=*]
    \item \textbf{Traditional  approach:} This includes Markov models~\cite{gambs2012next} that fit separate transition matrices for different datasets.
    \item \textbf{Deep mobility models:} We include LSTM~\cite{kong2018hst}, Transformer~\cite{vaswani2017attention}, DeepMove~\cite{deepmove}, TrajBert~\cite{si2023trajbert}, and CLET~\cite{CLET} as representative baselines. For each dataset, we train a separate model to ensure fair comparison under the same training conditions.
    \item \textbf{Fedarated learning models:} We evaluate against PMF~\cite{feng2020pmf} and LightTR~\cite{10598171LightTR}, which leverage federated learning frameworks for human mobility prediction while maintaining data decentralization. We apply their federated learning methods to train our model.
    \item \textbf{Joint models with privacy protection:} These methods enable continual learning without accessing previously seen data. Specifically, we consider two variants of our model: \textbf{MoveGCL (FullTune)} unfreezes all experts and routers in the MoE Transformer while keeping the rest of the model frozen, and fine-tunes using only the new city's data. \textbf{MoveGCL (ExpertTune)} incrementally adds one new expert per layer in the MoE Transformer, unfreezes all experts and routers, and fine-tunes on the new city's data while keeping all other parameters fixed.
\end{itemize}

\subsection{Parameter Settings}\label{sec:para}
The key parameters of our framework fall into three main categories.
\begin{itemize}[leftmargin=*]
    \item For the model architecture, we set the temporal embedding dimension to 48. In the mobility encoder, the embedding dimensions of $d_{\mathrm{jump}}$ and $t_{\mathrm{wait}}$ are both 128, the embedding dimensions of $r_{\mathrm{gyr}}$ and location entropy $H_{\mathrm{loc}}$ are 64, and the embedding dimension of $\mathrm{ID}_{\mathrm{city}}$ is 32, with the self‐attention modules for both $d_{\mathrm{jump}}$ and $t_{\mathrm{wait}}$ using a hidden dimension of 64. In the trajectory location encoder and the city location encoder, the embedding dimensions of \(\phi_{\text{POI}}\), \(\phi_{\text{hot}}\), and \(\phi_{\text{lat-lon}}\) are  256, 128, 128, respectively. The hidden dimension of each MoE transformer block is set to 512. The initial number of experts in each MoE transformer block is 4, and the model comprises 6 layers of MoE transformer blocks.
    \item For the base model training phase, the initial model is obtained by training on three cities' datasets. During this phase, we use a batch size of 16 and train for 30 epochs. The initial learning rate is set to \(1.2\times10^{-5}\), and the learning rate decays in a stepwise fashion during training.
    \item For generative continual learning, whenever we introduce a new dataset (i.e., a new city), we add one new expert to every MoE transformer block. The initial learning rate for this phase is \(1.2\times10^{-4}\), the batch size is 128, and training also runs for 30 epochs, with the learning rate decaying stepwise throughout. The generative coefficient \(\alpha\) is set to 20\%. The balance coefficient \(\lambda\) for \(\mathcal{L}_{\text{total}}\) is set to 1.
\end{itemize}

\subsection{Evaluation Metrics}

We adopt top-\(k\) accuracy as our evaluation metric, defined as
\begin{equation} 
\mathrm{acc}@k \;=\; \frac{1}{N} \sum_{i=1}^{N} \mathbf{1}\bigl(x_i \in f_k(x_i)\bigr)\,,
\end{equation} 
where \(N\) is the total number of samples, \(x_i\) is the ground-truth label for the \(i\)\textsuperscript{th} sample, \(f_k(x_i)\) denotes the set of the model’s top-\(k\) predicted labels for sample \(i\), and \(\mathbf{1}(\cdot)\) is the indicator function that equals 1 if its argument is true and 0 otherwise. We report on \(\mathrm{acc}@1\) and \(\mathrm{acc}@3\) to assess the performance of the model.

\section{Related Work}\label{sec:related}

\paragraph{\textbf{Mobility Data}}
Mobility data includes aggregated flows and individual trajectories~\cite{barbosa2018human,yuan2025learning,zhang2025noise}. Aggregated flows are relatively easier to obtain and have been widely used in urban analytics~\cite{rong2024interdisciplinary,zhang2017deep}. However, individual-level mobility data remain fragmented due to privacy concerns and institutional data silos~\cite{kong2023mobility,yuan2025learning}. Real-world trajectory datasets, such as GeoLife~\cite{zheng2011geolife}, T-Drive~\cite{yuan2010t}, NYC Taxi~\cite{nyc_tlc_trip_data}, and Foursquare~\cite{yang2019revisiting}, often have limited city coverage, short time spans, and sparse sampling. Some global-scale open datasets, such as the one used in UniTraj, have been introduced, but they suffer from low spatial-temporal resolution and inconsistent data quality.
With the advancement of generative AI, synthetic mobility datasets have emerged, such as SynMob~\cite{zhu2023synmob}, YJMob100K~\cite{yabe2024yjmob100k} and WorldMove~\cite{yuan2025worldmove}. However, the quality of synthetic data still falls short compared to real-world trajectories, particularly in terms of behavioral diversity, temporal continuity, and semantic consistency. In practice, access to real trajectory datasets typically requires signing NDAs, and most published studies do not release the mobility datasets they use due to privacy and legal restrictions~\cite{schlapfer2021universal}.

\paragraph{\textbf{Mobility Foundation Models}}
Inspired by the success of foundation models in NLP and CV, recent efforts have explored pre-trained models for urban and mobility domains~\cite{yuan2024urbandit, chai2025mobiworld, han2025trajmoe, zhou2024urban,zhang2024urban,choudhury2024towards}. Early studies primarily focused on aggregated mobility data, leveraging mobility flows across cities to build unified spatio-temporal representations, and have demonstrated strong zero-shot transfer capabilities~\cite{li2024urbangpt,yuan2024uniflow,yuan2024unist,li2024opencity}.
In contrast, individual-level mobility foundation models are less developed. Researchers have explored multi-scale mobility modeling~\cite{yuan2025learning,zhang2025noise,long2024universal}, aiming to capture both micro-level behaviors and macro-level patterns essential for generalization.  Attempts such as UniTraj~\cite{zhu2024unitraj}, TrajBert~\cite{si2023trajbert},  and TrajFM~\cite{lin2024trajfm} have explored learning from open trajectory datasets, but these datasets often consist of short-term or non-representative mobility traces that do not reflect regular human movement patterns. As a result, current models struggle to capture the full complexity and diversity of real-world individual mobility. Recently, LLMs have also widely utilized in generating human mobility~\cite{shao2024beyond,gong2024mobility,ju2025trajllm,feng2024agentmove,jiawei2024large}, but the gap between natural language and trajectory data suggests that mobility still requires native foundation models, which can later be aligned with LLMs to bridge symbolic reasoning and physical behavior modeling.

\paragraph{\textbf{Continual Learning}}
Continual learning~\cite{kim2022theoretical,wang2024comprehensive,wang2024comprehensive}, also known as lifelong learning, aims to enable models to learn from a sequence of tasks or data streams without forgetting previously acquired knowledge~\cite{yoon2021federated}. A central challenge in continual learning is catastrophic forgetting~\cite{li2019learn,wickramasinghe2023continual}, where the model's performance on earlier tasks degrades as it learns new ones.
Continual learning methods are typically categorized into three main types. Regularization-based methods introduce constraints on parameter updates to preserve important knowledge from earlier tasks.
Replay-based methods mitigate forgetting by either storing a subset of previous data (experience replay) or generating pseudo-data (generative replay) to simulate past learning. Parameter isolation methods allocate different parts of the model to different tasks, using techniques such as dynamic networks or task-specific masking to reduce interference between tasks.

\section{Additional Results}

\begin{table*}[t!]
\caption{Performance comparison across cities and Acc@k metrics for evaluating order invariance.}
\label{tbl:order}
\vspace{-3mm}
\centering
\resizebox{\textwidth}{!}{
\begin{tabular}{c cc cc cc cc cc cc}
\hline
\textbf{Model} & \multicolumn{2}{c}{\textbf{Atlanta}} & \multicolumn{2}{c}{\textbf{Chicago}} & \multicolumn{2}{c}{\textbf{Los Angeles}} & \multicolumn{2}{c}{\textbf{New York}} & \multicolumn{2}{c}{\textbf{Seattle}} & \multicolumn{2}{c}{\textbf{Washington D.C}} \\
\cmidrule(lr){2-3} \cmidrule(lr){4-5} \cmidrule(lr){6-7} \cmidrule(lr){8-9} \cmidrule(lr){10-11} \cmidrule(lr){12-13}
 & Acc@1 & Acc@3 & Acc@1 & Acc@3 & Acc@1 & Acc@3 & Acc@1 & Acc@3 & Acc@1 & Acc@3 & Acc@1 & Acc@3\\
\hline
WSC$\to$A$\to$L$\to$N     &0.282 & 0.421 & 0.197 & 0.306 & 0.157 & 0.254 & 0.206 & 0.328 & 0.324 & 0.478 & 0.273 & 0.413 \\
AWN$\to$L$\to$S$\to$C      &0.284 & 0.423 & 0.197 & 0.308 & 0.150 & 0.246 & 0.200 & 0.326 & 0.317 & 0.469 & 0.265 & 0.407 \\
WAL$\to$N$\to$S$\to$C     &0.285 & 0.426 & 0.194 & 0.304 & 0.151 & 0.245 & 0.188 & 0.306 & 0.321 & 0.472 & 0.267 & 0.407 \\
Joint Training            &0.288 & 0.428 & 0.192 & 0.302 & 0.156 & 0.250 & 0.198 & 0.320 & 0.322 & 0.475 & 0.270 & 0.410 \\
\hline
\end{tabular}
}
\end{table*}

\subsection{Order Invariance in Continual Learning}\label{sec:order}
To assess the robustness of MoveGCL in real-world deployment scenarios, we investigate its sensitivity to the order in which data from different cities is introduced during continual learning. In practice, the arrival of mobility data is often dictated by external factors such as data access regulations, infrastructure development cycles, or institutional collaborations. As a result, foundation models intended for long-term, large-scale deployment must remain robust to such variations in data sequencing.

We simulate this scenario by reversing the order of datasets used in the continual learning phase. As shown in Table~\ref{tbl:order}, the performance of MoveGCL remains remarkably consistent, with the vast majority of metrics showing deviations of less than 5\% across original and reversed sequences. This empirical finding confirms the order-invariant learning behavior of our model, demonstrating that MoveGCL can integrate new city data without disrupting previously acquired knowledge, even when the order of exposure varies significantly. 
This property is especially crucial for building scalable and unified mobility foundation models, which must support progressive, privacy-preserving knowledge accumulation in non-i.i.d. settings where data arrives incrementally and asynchronously. Order robustness is thus a key enabler for deploying foundation models that can continually evolve while ensuring stable performance and generalization across diverse urban contexts.

\subsection{Experimental Settings of Privacy Evaluation}~\label{sec:set_privacy}

\paragraph{\textbf{Uniqueness Testing}}
We randomly extract a subset of trajectories from the training set and use an autoregressive process to generate new trajectories conditioned on each sampled trajectory. We then compute the pairwise similarity between each original sample and its corresponding generated trajectory.If the lengths of two trajectories are different, the similarity score is defined as 0. If they are of equal length, the similarity score is calculated as the proportion of positions where both the timestamp and location ID exactly match. For each generated trajectory, we compute its similarity score with the top-1, top-3, and top-5 most similar real trajectories.

\paragraph{\textbf{Membership Inference Attack}}
Following the experimental setup in~\cite{lin2020doppelganger, shokri2017membership}, we use the similarity between generated trajectories and their corresponding real input trajectories as the classification feature. For each trajectory \(x\), MoveGCL generates \(\tilde{x}\) autoregressively conditioned on \(x\), and we compute a similarity score \(s(x, \tilde{x})\) to form the input to the classifier. The classifier is then tasked with determining whether \(x\) was included in the model’s training set. Positive samples consist of real‐world trajectories that were used during training, while negative samples are real trajectories from the same city that were held out. We evaluate the attack success rate, defined as the proportion of samples for which the classifier correctly infers membership status. We employ three widely used classification algorithms: Logistic Regression (LR), Support Vector Machine (SVM) and Random Forest (RF).

\paragraph{\textbf{Differential Privacy}}
For any pair of datasets \(D\) and \(D'\) that differ by only a small number of training trajectories, a model \(M\) is said to satisfy \((\varepsilon, \delta)\)-differential privacy if the following condition holds:
\begin{equation}
\mathbb{P}\bigl[M(z;\,D)=z\bigr] \;\le\; e^{\varepsilon}\,\mathbb{P}\bigl[M(z;\,D')=z\bigr] \;+\;\delta,
\end{equation}
where \(\mathbb{P}\bigl[M(z;\,D)=z\bigr]\) denotes the probability of observing output \(z\) when the model is trained on dataset \(D\), and \(\mathbb{P}\bigl[M(z;\,D')=z\bigr]\) is defined analogously for dataset \(D'\). Smaller values of \(\varepsilon\) and \(\delta\) imply stronger privacy guarantees, since the model’s output distribution becomes less dependent on any single trajectory.

In our experiment, we randomly select a subset of trajectories and consider two training scenarios: one in which this subset is included in the training data (\(D\)), and one in which it is excluded (\(D'\)). For each scenario, we train \(M\) on the corresponding dataset and then use each selected trajectory as a conditioning input to generate multiple synthetic trajectories. We compute a similarity score between each generated trajectory and its original conditioning trajectory. These similarity scores are modeled as two Gaussian distributions corresponding to \(\mathbb{P}\bigl[M(z;\,D)\bigr]\) and \(\mathbb{P}\bigl[M(z;\,D')\bigr]\), respectively. Finally, we estimate the privacy‐budget parameters \(\varepsilon\) from these distributions.

\subsection{Supplementary Figures}

\begin{figure*}[h]
  \centering
  \includegraphics[width=0.6\textwidth]{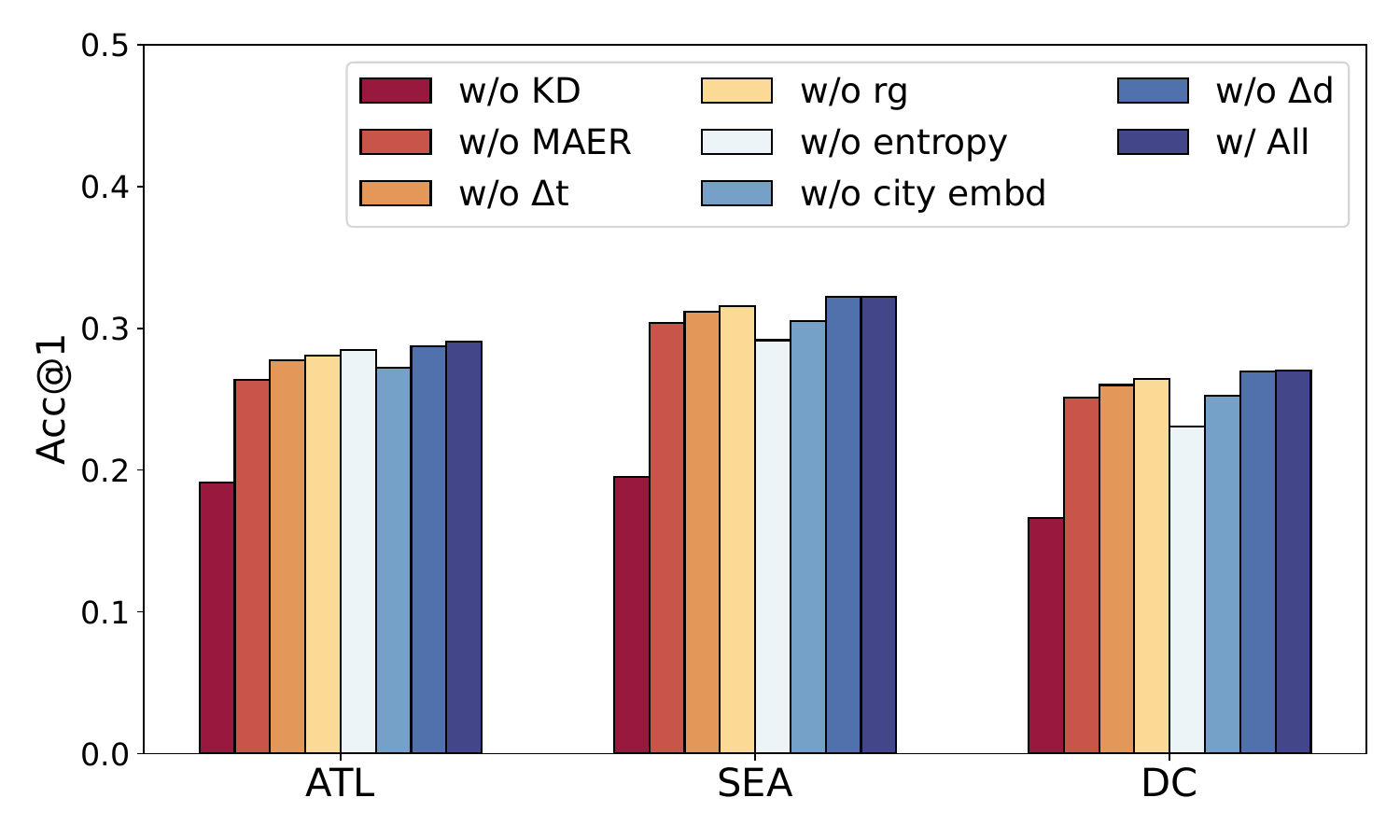}
  \caption{Ablation study. “w/o KD” denotes removal of the knowledge distillation loss; “w/o MAER” denotes removal of mobility feature from MoE transformer’s router inputs.} \label{fig:Ablation_Studies}
\end{figure*}

\begin{figure*}[h]
  \centering
  \includegraphics[width=0.8\textwidth]{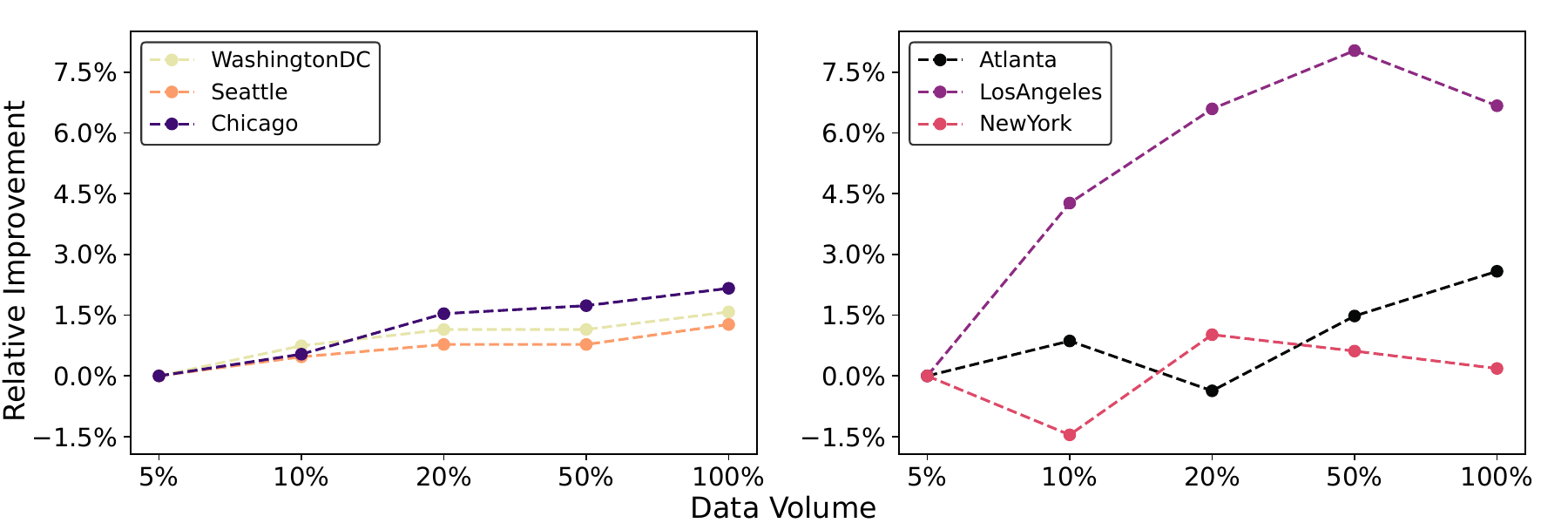}
  \caption{Acc@1 changes at different generated data ratios ($\alpha$), relative to $\alpha = 5\%$.} \label{fig:data_scale}
\end{figure*}

\end{document}